\documentclass[conference]{IEEEtran}
\IEEEoverridecommandlockouts
\usepackage{cite}
\usepackage{amsmath,amssymb,amsfonts}
\usepackage{algorithmic}
\usepackage{graphicx}
\usepackage{textcomp}
\usepackage{xcolor}

\usepackage{url}
\usepackage{booktabs}
\usepackage{tcolorbox} 
\usepackage{tabularx}      
\usepackage{adjustbox}
\usepackage{multirow}   
\usepackage{makecell}

\def\BibTeX{{\rm B\kern-.05em{\sc i\kern-.025em b}\kern-.08em
    T\kern-.1667em\lower.7ex\hbox{E}\kern-.125emX}}

\makeatletter
\newcommand{\linebreakand}{%
  \end{@IEEEauthorhalign}
  \hfill\mbox{}\par
  \mbox{}\hfill\begin{@IEEEauthorhalign}}
\makeatother

\begin{document}

\graphicspath{{Figures/}}

\title{ProIQA: A Process-Based Framework for Fine-Grained Math Item Quality Assessment\\
}



\author{
\IEEEauthorblockN{Junkai Tong}
\IEEEauthorblockA{\textit{East China Normal University}\\
Shanghai, China \\
51285901006@stu.ecnu.edu.cn}
\and
\IEEEauthorblockN{Mingjia Li}
\IEEEauthorblockA{\textit{East China Normal University}\\
Shanghai, China \\
limj@stu.ecnu.edu.cn}
\and
\IEEEauthorblockN{Haoran Chen}
\IEEEauthorblockA{\textit{East China Normal University}\\
Shanghai, China \\
10235102510@stu.ecnu.edu.cn}
\and
\IEEEauthorblockN{Yaoyu Jiang}
\IEEEauthorblockA{\textit{East China Normal University}\\
Shanghai, China \\
10235102489@stu.ecnu.edu.cn}
\linebreakand
\IEEEauthorblockN{Hanjie Ge}
\IEEEauthorblockA{\textit{East China Normal University}\\
Shanghai, China \\
51285901122@stu.ecnu.edu.cn}
\and
\IEEEauthorblockN{Yixuan Wang}
\IEEEauthorblockA{\textit{East China Normal University}\\
Shanghai, China \\
yxwang@stu.ecnu.edu.cn}
\and
\IEEEauthorblockN{Hong Qian\textsuperscript{*}}
\IEEEauthorblockA{\textit{East China Normal University}, Shanghai, China \\
\textit{Shanghai Innovation Institute}, Shanghai, China \\
hqian@cs.ecnu.edu.cn}
}

\maketitle

\begin{abstract}
Automatic Item Generation (AIG) is pivotal for personalized education, yet guaranteeing the pedagogical value of generated items remains a bottleneck. Existing Item Quality Assessment (IQA) methods typically rely on unscalable manual reviews or shallow stem-based metrics, failing to capture the reasoning process required for mathematical problem-solving. To bridge this gap, this paper proposes Process-based Item Quality Assessment (ProIQA), a process-aware framework for fine-grained quality assessment of math items. We first formulate IQA across three heterogeneous dimensions, including knowledge concepts, difficulty, and disciplinary competencies, under a unified process-aware perspective. Based on this formulation, we construct a process-enhanced IQA resource by augmenting original item data with structured reasoning trees derived from raw solutions. Technically, ProIQA leverages Large Language Modelsto construct hierarchical reasoning trees and employs Graph Neural Networks (GNN) to encode their topological dependencies and procedural semantics. The resulting solving representation is fused with stem semantics through a dual-view (``Stem + Solving'') architecture, enabling comprehensive assessment across learning objectives. Extensive experiments on K12 mathematical datasets show that ProIQA effectively captures process-oriented features, offering a scalable data-driven solution for evaluating AIG outputs in intelligent education systems.

\end{abstract}


\section{Introduction}

In the rapidly evolving landscape of intelligent education, Automatic Item Generation (AIG) has emerged as a transformative technology, critical for promoting educational equity and enabling personalized learning~\cite{AIG}. By leveraging algorithms to generate diverse and high-quality assessment items, AIG facilitates the creation of tailored exercises that target specific student knowledge weaknesses, thereby supporting the implementation of adaptive learning systems and intelligent diagnostic services~\cite{CAT}. Despite these advancements in generation capabilities, Item Quality Assessment (IQA) remains a central challenge. Ensuring that generated items possess appropriate pedagogical value, such as suitable difficulty, knowledge coverage, and clarity is essential before their deployment. 
However, the lack of a reliable, fine-grained evaluation framework limits AIG's downstream utility~\cite{AQG}.

Currently, approaches to IQA generally fall into two categories, both of which face significant limitations. On one hand, traditional IQA relies heavily on manual review by subject matter experts or small-scale pilot study with student groups~\cite{e-rater_V.2}, which is labor-intensive, time-consuming, and prone to subjective inconsistencies, making it fundamentally unscalable for the massive, rapidly growing libraries generated by AIG algorithms. On the other hand, while recent automated evaluation techniques have attempted to address these scalability issues, they remain limited in scope and depth. Most existing automated approaches focus predominantly on Multiple Choice Questions (MCQs), utilizing static rubrics or shallow statistical metrics (e.g., readability scores) to estimate quality~\cite{AQG}. Even when targeting specific dimensions such as knowledge coverage or difficulty, these methods usually assess items mainly from the surface text of the stem, limiting their ability to handle complex open-ended mathematical problems.

To overcome these barriers, it is necessary to reconsider the cognitive nature of quality assessment. When human experts evaluate a math problem, they not only scrutinize the stem text but also simulate the solution process to identify which concepts are invoked, how many reasoning steps are required, and what type of mathematical thinking is involved~\cite{rule_space}. These process-level signals are closely related to multiple dimensions of item quality, including knowledge concepts, difficulty, and disciplinary competencies. However, standard stem-only methods ignore this deep solving path, which explains why they often fail to match expert judgment. Therefore, we propose to incorporate solution process representation into the assessment loop, constructing a \textbf{dual-view (``Stem + Solving'')} framework that combines surface-level item stem (question statement) with deep-level reasoning processes.

Implementing this process-oriented approach introduces two technical challenges. The first challenge lies in \textbf{representation construction}: unlike unstructured narrative text, mathematical reasoning is strictly logical and hierarchical. Transforming raw item text into a structured representation that explicitly captures the reasoning flow rather than a linear sequence of tokens is non-trivial. The second challenge is \textbf{semantic extraction}: even with a structured representation, effectively encoding the procedural semantics within logical steps and their dependencies requires advanced modeling techniques beyond standard text encoders. Existing methods often struggle to bridge the gap between the surface-level problem statement and the deep-level cognitive effort to solve it.

To address the aforementioned two challenges, this paper proposes \textbf{Process-based Item Quality Assessment (ProIQA)}, which implements the solving view through structured reasoning trees. Leveraging the reasoning capabilities of LLMs, we automate the construction of reasoning trees, decomposing complex problems into hierarchical sub-steps and augmenting original item data with explicit process structures. We employ GNNs with multi-layer architectures to encode the topological dependencies within these trees. By fusing the surface-level semantics from the item stem with the deep-level procedural semantics from the reasoning tree through dual-view representation learning, ProIQA provides a \textbf{unified process-aware framework} for fine-grained item quality assessment. In contrast to prior stem-based methods, this framework does not force heterogeneous dimensions into a single label space. Instead, knowledge concepts, difficulty, and disciplinary competencies share one process-enriched representation while preserving task-specific learning objectives.

The main contribution of this work is summarized as follows:
\begin{itemize}

\item \textbf{Unified Process-Aware Formulation and Process-Enhanced IQA Resource:} We formulate IQA across knowledge concepts, difficulty, and disciplinary competencies under a unified process-aware perspective. Based on this formulation, we construct a process-enhanced K12 mathematical IQA resource by augmenting item data with structured reasoning trees derived from raw solutions, providing reusable hierarchical and procedural information for educational data mining and IQA research.

\item \textbf{Reasoning-Tree Based Dual-View Representation Learning:} 
ProIQA combines stem semantics with solving semantics, using LLMs to build hierarchical reasoning trees, GNNs to encode their topological dependencies, and a dual-view fusion of both representations. This supports objectives such as pairwise difficulty ranking and multi-label concept mapping.

\item \textbf{State-of-the-Art Performance and Validation:} Extensive experiments across diverse K12 mathematical datasets show the strong performance of ProIQA over both supervised baselines and advanced LLMs. Our method achieves average performance gains of $7.5\%$ in concept assessment, $6.3\%$ in difficulty estimation, and $19.5\%$ in competency assessment compared to the second-best algorithms. These results, supported by ablation studies and stability analysis, validate the effectiveness of integrating process-oriented representations into item quality assessment.
\end{itemize}

The paper is organized as follows. Section 2 reviews related work. Section 3 formulates the assessment tasks, while Section 4 details the ProIQA framework. Section 5 presents the experimental results, followed by conclusions in Section 6.

\section{Related Work}

\subsection{Automated Item Generation}

The evolution of AIG has shifted from rigid template-based methods to flexible LLMs, enabling the scalable creation of complex mathematical items~\cite{2017learning_to_ask,NQG,Advances_in_NQG,PoT}. However, quality assurance remains a critical bottleneck~\cite{AQG}. Generated content is prone to hallucinations and logical flaws~\cite{Survey_on_LLMEducation}. Current evaluation paradigms largely rely on labor-intensive human review~\cite{ASP} or shallow automated metrics (e.g., N-gram, ROUGE-L, readability scores)~\cite{AEQG,Auto-Assessment,MWP,MWP-OPT}. These approaches struggle to capture fundamental educational properties, such as specific knowledge coverage and difficulty levels, highlighting growing demand for a fine-grained, scalable IQA framework.

\subsection{Item Quality Assessment}
Existing IQA approaches generally fall into two categories: \textit{response-based} and \textit{content-based}. Response-based methods, such as pilot testing via IRT~\cite{IRT} or recent LLM-simulated student profiles~\cite{students_on_MCQs, Student_Simulations_on_MWP}, or cognitive diagnosis~\cite{FineCD, hypergraph_cd,qccdm}, suffer from either low scalability or ``fidelity gap'' between simulated and real student behaviors. Content-based automated methods have evolved from handcrafted features~\cite{C-test, deane2019scenario,difficulty_FOL} to deep learning models for difficulty prediction~\cite{prediction_for_difficulty,DKFM, MCQ_in_Medicalexam,R2DE,transformers_difficulty_mcq,KGNN-ADP,Neural_Networks_MCQ}. A systematic review can be found in~\cite{review_automatic_approaches}.

Despite their advancements, existing automated methods share two key limitations. First, most studies focus exclusively on MCQ difficulty prediction, neglecting other critical dimensions such as concept coverage and disciplinary competencies. Second, these models typically perform \textbf{surface-level analysis} on \textbf{item stems}, treating math problems as standard text classification tasks and largely overlooking the specific \textbf{solution logic} and multi-step reasoning required to solve them. While recent work has begun to utilize auxiliary reasoning trajectories, how to effectively integrate such process-level information into quality prediction remains an open question. 
Existing graph or transformer methods, such as QuesImNet~\cite{quesimnet}, represent questions via text or knowledge graphs, but none model the solution process captured by ProIQA's reasoning trees.


To bridge this gap, ProIQA shifts from ``Stem-based'' to ``Process-based'' assessment. By leveraging LLMs to reconstruct the explicit reasoning process and fuse it with item information, our framework captures the cognitive depth and logical dependencies of an item, enabling a holistic and automated evaluation of concepts, difficulty, and competencies.

\begin{figure*}[t]
    \centering
    \includegraphics[width=0.85\textwidth]{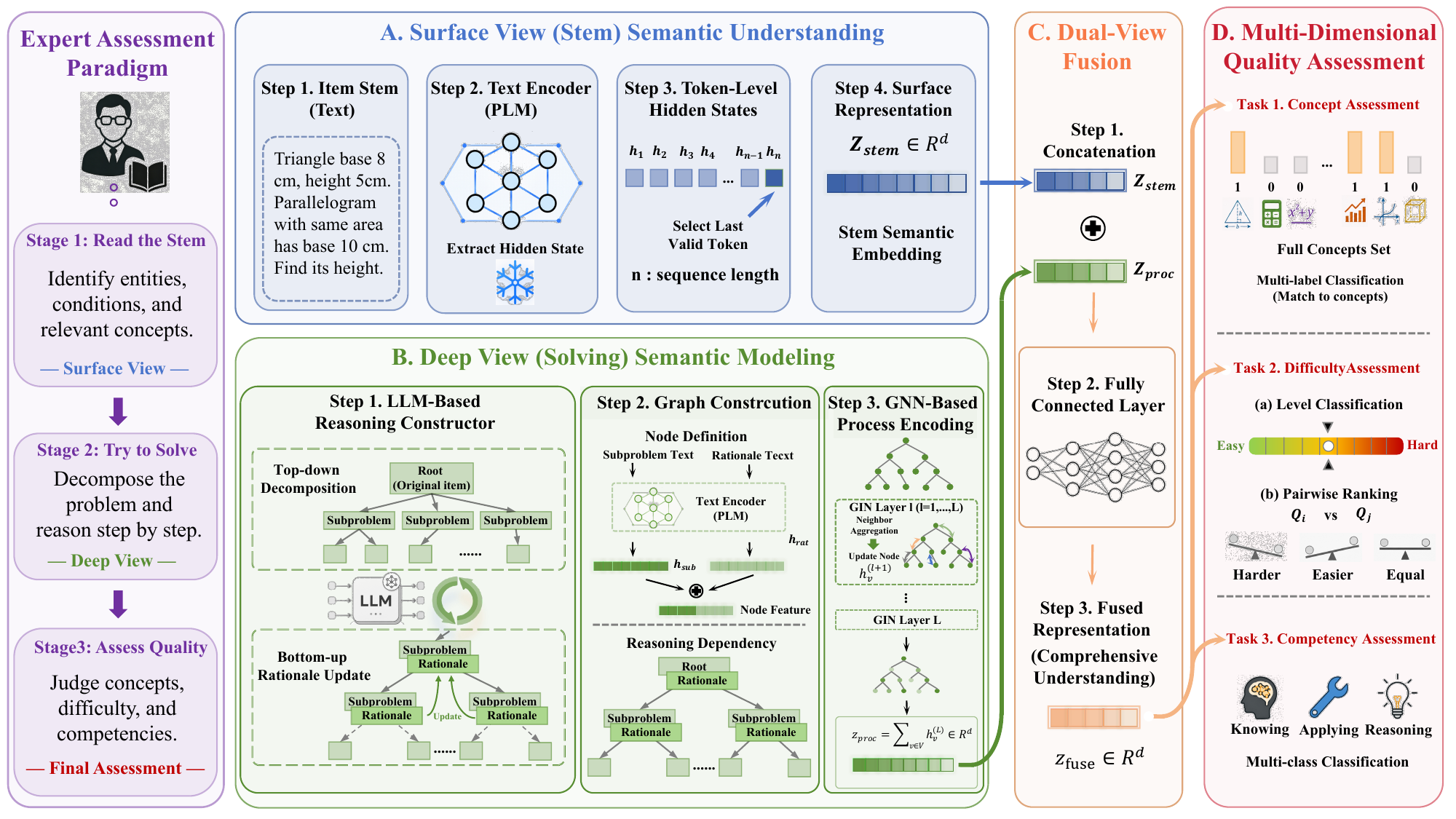}
    \caption{Overall architecture of ProIQA. Inspired by the human expert assessment paradigm, ProIQA performs dual-view representation learning by combining surface stem semantics with solving-process semantics derived from structured reasoning trees. The fused representation supports multi-dimensional item quality assessment across concept, difficulty, and competency dimensions.}

    \label{fig:framework}
\end{figure*}

\section{Problem Formulation of IQA}



In this section, we define the input data and formalize the IQA tasks within the ProIQA framework. The primary goal of these tasks is to serve as a robust quality assurance mechanism for AIG systems, ensuring that generated items align with pedagogical requirements in terms of knowledge coverage, difficulty calibration, and cognitive depth.


\subsection{Notations}
Let $\mathcal{Q} = \{Q_1, Q_2, \dots, Q_N\}$ denote the set of generated math items to be evaluated. Each item $Q_i$ is represented by its textual stem. The objective is to map each $Q_i$ to labels in three specific dimensions:
\begin{itemize}
    \item \textbf{Concepts:} Let $\mathcal{C} = \{C_1, C_2, \dots, C_K\}$ be the universal set of knowledge concepts.
    \item \textbf{Difficulty:} Let $\mathcal{D}$ represent the difficulty metric, which can be viewed as continuous values $d \in \mathbb{R}$ or discrete levels $l \in \{1, \dots, D\}$.
    \item \textbf{Competencies:} Let $\mathcal{P} = \{p_k, p_a, p_r\}$ represent the set of disciplinary competency categories (knowing, applying, reasoning).
\end{itemize}

\subsection{Assessment Tasks}

While the educational measurement literature has long recognized the multi-dimensional nature of item quality—encompassing knowledge coverage, difficulty, and cognitive competencies—these dimensions have largely been treated in isolation within the machine learning community. Prior automated IQA studies typically address a single dimension (most commonly difficulty prediction) using generic classification or regression formulations, lacking a principled framework that aligns the learning objective with the intrinsic structure of each assessment dimension. For instance, concept coverage is inherently a sparse matching problem between items and a dynamic knowledge taxonomy, yet it is often reduced to fixed multi-label classification. Similarly, difficulty assessment in adaptive testing scenarios benefits more from relative pairwise comparisons than from absolute score regression, but this distinction is rarely formalized. In this section, we present a unified problem formulation that systematically defines three core IQA tasks, each with a tailored learning objective designed to reflect the authentic requirements of educational quality assurance. This formalization not only provides a consistent evaluation protocol for the proposed ProIQA framework, but also establishes a reusable task definition template for future IQA research.

\subsubsection{Dimension 1 (Concept)} In AIG systems, items are typically generated based on a specific query or syllabus constraint (e.g., ``Generate a question about \textit{Geometry}''). A critical quality check is to verify whether the generated item actually covers the intended concept and does not drift into unrelated or out-of-syllabus topics. This alignment verification ensures the validity of the item for targeted practice.
\par
\noindent \textbf{Subtask 1:} We formalize concept assessment as an \textbf{Item-Concept Matching} task. Rather than predicting a fixed label vector, we treat this as a binary classification problem to determine the relevance between an item and a concept query. Given an item $Q_i$ and a candidate concept $C_k$, the model learns a mapping function $f_{con}$ : $ \hat{y}_{i,k} = f_{con}(Q_i, C_k) \in \{0, 1\} , $ where $\hat{y}_{i,k}=1$ indicates that $Q_i$ successfully aligns with concept $C_k$. This formulation allows for flexible validation against dynamic concept lists in evolving educational standards.

\subsubsection{Dimension 2 (Difficulty)} A key advantage of AIG is supporting adaptive learning by providing items match a student's current proficiency level~\cite{CAT}. To achieve this, generated items need to be calibrated. Difficulty assessment serves to (1) filter items into general difficulty tiers for test paper balancing~\cite{PCAT}, and (2) predict difficulty parameters for Item Response Theory (IRT)~\cite{IRT} based recommendation algorithm.
\par
\noindent \textbf{Subtask 2:} We formulate difficulty assessment as a hierarchical task combining broad classification with fine-grained pairwise comparison:
\begin{itemize}
    \item \textbf{Level Classification:} To categorize items into general tiers (e.g., Easy, Medium, Hard): $\hat{l}_i = f_{cls}(Q_i) \in \{1, \dots, D\} .$  
    
    \item \textbf{Pairwise Relative Comparison:} Instead of regressing absolute values, we formulate fine-grained difficulty estimation as a pairwise ranking problem. Let $d_i$ denote the ground-truth difficulty parameter of item $Q_i$, estimated via IRT~\cite{IRT} from student response logs. For each item pair $(Q_i, Q_j)$, we define a ternary label $y_{ij} = \mathrm{sign}(d_i - d_j) \cdot \mathbf{1}\left[\,|d_i - d_j| > \epsilon\,\right]$ with margin $\epsilon = 0.5$, and learn a comparator $f_{\mathrm{pair}}$ to predict it: $\hat{y}_{ij} = f_{\mathrm{pair}}(Q_i, Q_j) \in \{-1,0,+1\}$. By minimizing the prediction error on these pairs, the model captures subtle semantic features contributing to relative difficulty differences, enabling precise item ranking for adaptive testing.
\end{itemize}

\subsubsection{Dimension 3 (Competency)} A common criticism of AIG is the tendency to generate shallow questions that test rote memory. Ensuring high-quality education, it is essential to distinguish whether a generated item merely requires recalling facts (\textit{Knowing}) or demands logic and critical thinking (\textit{Reasoning}). This assessment ensures generated library maintains a healthy distribution of cognitive depths suited for cultivating higher-order thinking skills.
\par
\noindent \textbf{Subtask 3:} Adopting the TIMSS framework~\cite{TIMSS2019}, we formalize competency assessment as a \textbf{Multi-class Classification} task. The model assigns each item to a unique cognitive domain that represents its primary requirement: $\hat{p}_i = f_{comp}(Q_i) \in \{\text{Knowing, Applying, Reasoning}\}\,$. 
    
This classification acts as a filter to guarantee the AIG delivers items with sufficient pedagogical value and complexity.

\section{The Proposed ProIQA}

The overall framework of ProIQA is illustrated in Figure~\ref{fig:framework}. Our approach simulates the expert evaluation paradigm by modeling the item solution process. It consists of three components: (1) \textbf{Reasoning Tree Construction}, which transforms raw item stems into structured reasoning trees; (2) \textbf{Dual-View Representation Learning}, which encodes both surface-level textual semantics and deep-level procedural logic; and (3) \textbf{Multi-Dimensional Assessment}, which performs task-specific predictions for concepts, difficulty, and competencies.


\subsection{Reasoning Tree Construction}

Traditional stream-based solution generation often fails to capture hierarchical dependencies between logical steps essential for quality assessment. To address this, we construct a structured \textbf{Reasoning Tree} $T = (r, \mathcal{V}, \mathcal{E})$ for each item, where $r$ is root, $\mathcal{V}$ is node set, and $\mathcal{E}$ represents logical dependencies.

Inspired by the existing decomposition-solving paradigm~\cite{DeAR, ToT, Least-to-Most, Plan-and-Solve, Self-Refine}, we employ an LLM to recursively break down the item stem (root node $r$) into interconnected subproblems. This follows a top-down decomposition and bottom-up solution update mechanism. As shown in Figure \ref{fig:framework}, complex problems are decomposed into intermediate nodes (subproblems) until directly solvable. To ensure generalizability across various item types (e.g., Multiple Choice, Fill-in-the-Blank), we unify all items into a constructed-response format before tree generation. However, LLMs may produce reasoning trees of poor quality due to hallucinations or incoherent generation. To mitigate this, we integrate an automated quality assurance step into the pipeline. Specifically, we employ an LLM (e.g., Deepseek-R1) as a verifier to examine every generated reasoning tree for structural integrity, logical connectivity, and the mathematical correctness of each rationale before it is used for downstream training. We define \textit{verification accuracy} as the proportion of generated trees that pass this review.

Each node $v_i \in \mathcal{V}$ contains two semantic components: the \textit{subproblem text} ($S_{v_i}$) describing the immediate task, and the \textit{rationale text} ($R_{v_i}$) detailing the solution step of $S_{v_i}$. This structure exposes the reasoning trees needed for fine-grained assessment. Due to space constraints, we provide examples of the generated reasoning trees in our code repository.

\subsection{Dual-View Representation Learning}

\subsubsection{Surface View} We encode item stem through a frozen pre-trained language model (PLM) and take the hidden state of last valid token as the stem representation. This representation captures the surface-level semantics of item statement, including entities, quantities, and constraints stated in the item, yet carries no information about how item is actually solved. 

\subsubsection{Deep View} To complement the surface view, we extract solving semantics from the reasoning tree (Section IV-A). Node features are PLM-encoded from each node's subproblem and rationale, and a multi-layer GIN propagates along dependency edges to distill a global solving-level representation. This captures the hierarchical cognitive structure of the solution path that surface text alone misses. Details follow.

\noindent \textbf{Node Representation:} Existing methods often treat text and logic in isolation. We propose a collaborative encoding strategy. For each node $v_i$, we extract semantic features from both subproblem text $S_{v_i}$ and rationale text $R_{v_i}$ using PLM. We use the hidden state of last token as sentence embedding : $ \mathbf{h}_{v_i}^{sub} = \text{PLM}(S_{v_i}) \in \mathbb{R}^d\ $, $ \mathbf{h}_{v_i}^{rat} = \text{PLM}(R_{v_i}) \in \mathbb{R}^d $ .
These embeddings are concatenated to form the initial node representation $\mathbf{x}_{v_i}^{(0)}$, fusing task description with procedural reasoning: $\mathbf{x}_{v_i}^{(0)} = [\mathbf{h}_{v_i}^{sub} \oplus \mathbf{h}_{v_i}^{rat}] \in \mathbb{R}^{2d}$,
where ``$\oplus$'' denotes vector concatenation.

\noindent \textbf{Procedural Structure Encoding}
To capture the dependencies between reasoning steps, we employ a Graph Isomorphism Network (GIN)~\cite{GIN}. GIN is chosen for its superior discriminative power in distinguishing graph structures. The representation of node $v$ at the $l$-th layer is updated by aggregating features from its neighbors $\mathcal{N}(v)$:
$\mathbf{h}_v^{(l+1)} = \text{MLP}^{(l)} \left( (1+\epsilon^{(l)}) \mathbf{h}_v^{(l)} + \sum_{u \in \mathcal{N}(v)} \mathbf{h}_u^{(l)} \right)\,$
where $\epsilon$ is a learnable parameter and MLP denotes a Multi-Layer Perceptron. After $L$ layers of propagation, nodes integrate global structural information. To obtain the process-level representation $\mathbf{z}_{P}$ for the entire tree, we apply a readout function using summation pooling to preserve the magnitude of reasoning complexity:
$ \mathbf{z}_{P} = \mathbf{W}_{out} \cdot \left( \sum_{v \in \mathcal{V}} \mathbf{h}_v^{(L)} \right) + \mathbf{b}_{out}\,$.

\subsection{Multi-Dimensional Assessment}

Based on the procedural representation $\mathbf{z}_{P}$ and the static item stem representation $\mathbf{z}_{stem} = \text{PLM(Stem)}$, we tailor specific prediction heads for each quality dimension.

\paragraph{Concept Assessment.}
We treat concept assessment as a semantic matching task. Given a candidate concept $C_k$ with textual representation $\mathbf{z}_{C_k}$, we concatenate it with the item's dual-view representations. A binary classifier predicts the relevance probability $\hat{y}_{con}$ : $ \mathbf{z}_{con} = \mathbf{z}_{stem} \oplus \mathbf{z}_{P} \oplus \mathbf{z}_{C_k}\,$, $ \hat{y}_{con} = \mathbb{I} \left[ \sigma \left( \text{MLP}_{con}(\mathbf{z}_{con}) \right) \ge \tau \right]\,$.  
where $\sigma$ is the sigmoid function and $\tau$ is the decision threshold. This design allows the model to adapt to dynamic concept sets by encoding the concept text directly.

\paragraph{Difficulty Assessment.}
Difficulty is assessed through two complementary objectives using the fused item representation $\mathbf{z}_{diff} = \mathbf{z}_{stem} \oplus \mathbf{z}_{P}$:
\begin{itemize}
    \item \textbf{Regression:} To predict a precise difficulty value (or relative ranking score) $\hat{y}_{reg} \in \mathbb{R}$ : $ \hat{y}_{reg} = \text{Linear}(\text{MLP}_{reg}(\mathbf{z}_{diff}))\,.$
    \item \textbf{Classification:} To predict the difficulty level $\hat{y}_{cls} \in \{1, \dots, M\}$ : $        \hat{y}_{cls} = \text{Softmax}(\text{MLP}_{cls}(\mathbf{z}_{diff}))\,.$ 
\end{itemize}

\paragraph{Competency Assessment.}
Competency assessment relies heavily on the complexity of solution path. For datasets lacking item stems, we derive the procedural representation $\mathbf{z}_{P}$ directly from the solution text. The competency category $\hat{y}_{comp}$ (e.g., Knowing, Applying, Reasoning) is predicted via: $\hat{y}_{comp} = \text{Softmax}(\text{MLP}_{comp}(\mathbf{z}_{P}))\,.$

\section{Experiment}
We structure the experiments to answer the following research questions:
\begin{itemize}
\item \textbf{RQ1:} How effective is ProIQA in fine-grained item quality assessment across the three dimensions of concept identification, difficulty estimation, and competencies categorization compared to state-of-the-art baselines?
\item \textbf{RQ2:}  Does the incorporation of process-oriented reasoning representations contribute to performance improvements in concept and difficulty assessment? 
\item \textbf{RQ3:} How robust is ProIQA with respect to different hyperparameter settings?
\item \textbf{RQ4:} How does ProIQA jointly assess knowledge concepts, difficulty, and competencies through a case study in a unified setting?
\end{itemize}

In this section, we conducted extensive experiments across three tasks and their respective datasets to evaluate the performance of ProIQA. Multi-dimensionally annotated open-source datasets remain critically scarce in educational AI. Our selection of XES3G5M~\cite{XES3G5M} (elementary arithmetic), MATH-Algebra~\cite{MATH} (secondary symbolic reasoning), and TIMSS19/23 (international competency benchmarks) is deliberate in spanning the full cognitive spectrum of K12 mathematics, from concrete numerical manipulation to abstract multi-step deduction. The XES3G5M dataset is a large-scale student response dataset on mathematics for third-grade elementary students, collected from K12 online learning platform developed by TAL Education Group. This diversity across educational stages and assessment standards ensures that the evaluation reflects authentic K12 requirements rather than a single grade level or domain. The code is available at \url{https://github.com/qky7/ProIQA}.

\begin{table}[t]
\centering
\caption{Statistics of the K12 mathematical datasets used for the three IQA subtasks. ``--'' indicates attribute not apply.}

\label{tab:Statistics of all Datasets}
\begin{tabular}{w{c}{2cm}|w{c}{0.55cm}w{c}{0.7cm}|w{c}{0.5cm}w{c}{0.7cm}|w{c}{0.5cm}w{c}{0.5cm}}
\toprule
Task  & \multicolumn{2}{c|}{Concept} & \multicolumn{2}{c|}{Difficulty} & \multicolumn{2}{c}{Competency} \\
Dataset & {\fontsize{7pt}{8pt}\selectfont XES-500} & {\fontsize{7pt}{8pt}\selectfont XES-1600} & {\fontsize{7pt}{8pt}\selectfont Algebra} & {\fontsize{7pt}{8pt}\selectfont XES-1500} & {\fontsize{7pt}{8pt}\selectfont TIMS19} & {\fontsize{7pt}{8pt}\selectfont TIMS23} \\
\midrule
Number & 504   & 1,630  & 1,508  & 1,063  & 292   & 272 \\
concepts & 61    & 213   & —     & —     & —     & — \\
Competency & —     & —     & —     & —     & 3     & 3 \\
(avg) concepts & 1.95  & 1.98  & —     & —     & —     & — \\
(avg) difficulty & —     & —     & 3.21  & 0.55  & —     & — \\
(avg) stem length & 244   & 193   & 128   & 189   & —     & — \\
(avg) solution length & 134   & 107   & 341   & 103   & 38    & 40 \\
(avg) tree nodes & 5.27  & 4.68  & 4.31  & 4.55  & —     & — \\
(avg) tree length  & 1,631  & 1,259  & 1,229  & 1,161  & —     & — \\
\bottomrule
\end{tabular}%

\end{table}

\begin{table}[t]
\centering
\caption{Verification accuracy of the constructed reasoning trees across datasets.}
\label{tab:Reasoning Tree Quality}
\begin{tabular}{cc|cc|cc}
\toprule
\multicolumn{2}{c|}{Task} & \multicolumn{2}{c|}{(a) Concept} & \multicolumn{2}{c}{(b) Difficulty} \\
\multicolumn{2}{c|}{Dataset} & XES-500 & XES-1600 & Algebra & XES-1500 \\
\midrule
\multicolumn{2}{c|}{Accuracy} & 94.59\% & 90.38\% & 97.80\% & 91.46\% \\
\bottomrule

\end{tabular}%
\end{table}

\subsection{To RQ1: Multi-Dimensional Assessment Performance}

\paragraph{Dataset Preparation}
We curated six datasets spanning the three assessment tasks. Detailed statistics for all datasets are provided in Table~\ref{tab:Statistics of all Datasets}.

\begin{itemize}
\item \textbf{Concept assessment} : Due to computational constraints, we constructed two subsets, XES-500 (500 items) and XES-1600 (1600 items), by randomly sampling items from the original XES3G5M dataset. In XES3G5M,knowledge is annotated as hierarchical ``kc-routes'' (coarse-to-fine). We extracted the first two levels of routes as the target concepts for assessment.

\item \textbf{Difficulty assessment} : (1) Algebra: Constructed from algebra subset of MATH [39], where items are pre-labeled with difficulty (1--5). (2) XES-1500: A subset of 1500 items from XES3G5M. Since XES3G5M lacks authentic difficulty labels, we pre-estimated them using a two-parameter logistic (2PL) Item Response Theory (IRT) model~\cite{CAT} fitted to historical response logs. Note ProIQA does not require difficulty labels during processing. 

\item \textbf{Competency assessment} : We utilized the TIMSS
2019 and 2023 datasets (TIMSS19 \& TIMSS23), filtering
for K12 mathematical solution processes. These datasets cat-
egorize processes into three cognitive domains: Knowing,
Applying, and Reasoning.

\end{itemize}

\begin{table}[t]
\centering
\caption{Attributes of the LLMs used in zero-shot and fine-tuned benchmark settings. ``--'' indicates attribute unknown.}

\label{tab:StatisticsOfLLM}

\setlength{\tabcolsep}{2pt}
\begin{tabular}{cccccc}
\toprule
Model & Open  & Access & Parameters & Context & Type \\
\midrule
GPT-4o & $\times$     & API   & —     & 128K  & Zero-shot \\
Deepseek-R1 & $\checkmark$     & Weights & 671B  & 128K  & Zero-shot \\
Qwen3-235B & $\checkmark$     & Weights & 235B  & 32K   & Zero-shot \\
Qwen3-max & $\times$     & API   & —     & 256K  & Zero-shot \\
Gemini-2.5-pro & $\times$     & API   & —     & 1M    & Zero-shot \\
Llama-3.3-70B & $\checkmark$     & Weights & 70B   & 128K  & Zero-shot \\
Qwen3-8B & $\checkmark$     & Weights & 8B    & 32K   & Zero-shot / SFT \\
Llama-3.1-8B & $\checkmark$     & Weights & 8B    & 128K  & Zero-shot / SFT \\
Mistral-7B-Instruct-v0.3 & $\checkmark$     & Weights & 7B    & 32K   & Zero-shot / SFT \\
\bottomrule
\end{tabular}%

\end{table}

For the four datasets containing item stems (XES-500, XES-1600, Algebra, and XES-1500), we construct structured reasoning trees via the recursive decomposition procedure described in Section IV-A and validate them through LLM-based automated quality assurance. As reported in Table~\ref{tab:Reasoning Tree Quality}, the verification accuracy ranges from 90.38\% (XES-1600) to 97.80\% (Algebra) across all datasets, with an average success rate exceeding 90\%. These results indicate that the majority of generated trees correctly capture the topological dependencies among problem-solving steps, ensuring that the downstream GNN encoder consistently receives high-quality structured input. These trees capture the hierarchical decomposition of solution steps and, together with the original benchmarks, constitute a process-enhanced K12 mathematical IQA resource.

\paragraph{Experimental Setup}
For all three tasks, we adopted an 80\%--20\% train-test split with 5-fold cross-validation and reported the average results. All tasks share the same text embedding backbone (Qwen3-8B-Embedding), from which we extract the final hidden state (dim=4,096) as the semantic representation. Models were trained using the Adam optimizer. The task-specific configurations are as follows:
\begin{itemize}
   \item \textbf{Concept assessment} : The GNN was configured with 2 layers for XES-500 and 5 layers for XES-1600 (hidden dimension 512, dropout 0.4). The learning rate was set to $2e^{-5}$ with a max token length of 500.
   
   \item \textbf{Difficulty assessment} : The GNN was configured with 3 layers (hidden dimension 512, dropout 0.4) and a learning rate of $4e^{-5}$ (max token length 500). For XES-1500, we adopted a pairwise ranking approach, constructing item pairs $(Q_i, Q_j)$ and assigning labels $\{-1, 0, 1\}$ corresponding to $\{d_i < d_j, d_i \approx d_j, d_i > d_j\}$ based on a threshold of 0.5 difference in IRT difficulty parameters.
   
   \item \textbf{Competency assessment} : Since the TIMSS datasets provide solution texts but lack item stems, we bypassed both tree construction step and the GNN encoder, directly trained a 4-layer MLP classifier (lr=$1e^{-4}$) on the embedded solution text (max token length 150).
\end{itemize}

\paragraph{Baselines and Compared Methods}
ProIQA is compared with two categories of methods:

\begin{itemize}

   \item \textbf{LLM-based Zero-shot Methods}: We employ state-of-the-art LLMs in a zero-shot setting to benchmark the inherent capabilities of large-scale models. The lineup includes GPT-4o, Qwen3-235B, Gemini-2.5-Pro, DeepSeek-R1, Llama-3.3-70B, Qwen3-8B, Llama3.1-8B and Mistral-7B-Instruct-v0.3. These models were prompted to directly predict task labels given the item text (or solution text for the competency task). We provide detailed information about the LLM used in each task, as shown in Table~\ref{tab:StatisticsOfLLM}. Due to space constraints, all prompts used in this work are provided in our code repository.
   
   \item \textbf{Supervised Baselines} : These methods utilize the training data to update model parameters, ensuring fair comparison with ProIQA. (1) Supervised Fine-Tuning (SFT): To bridge the gap between supervised and zero-shot settings, we performed SFT on a suite of open-source LLMs at the 7B--8B scale using our training set. This serves as a strong generative baseline adapted to the specific task distribution. (2) T-IRT~\cite{transformers_difficulty_mcq}: Fine-tunes a pre-trained BERT model to extract deep contextualized semantic representations from the item stem. (3) R2DE~\cite{R2DE}: Represents item text using TF-IDF features to capture statistical keyword frequencies, followed by a Random Forest model to predict target attributes. For the competency assessment task, T-IRT and R2DE are not applicable due to the absence of item stems in the TIMSS datasets.

\end{itemize}

\paragraph{Evaluation Metrics}
Different metrics are adopted for each task to align with their specific learning objectives.
\begin{itemize}
   \item \textbf{Concept assessment} : Since concept assessment is a multi-label classification task with sparse labels, we utilize the F1-score as the primary metric.

   \item \textbf{Difficulty assessment} : We employ three metrics. Level-ACC measures the accuracy of the 5-way classification task, calculating the match rate between predicted and ground-truth levels. Pair-ACC measures the accuracy of the ternary pairwise comparison task, assessing the model's ability to correctly identify relative difficulty relationships (harder, easier, or tied). Weighted Adjacent Accuracy (WACC) is defined for 5-level classification in Algebra: standard ACC penalizes all errors equally, but a prediction off by one level (e.g., predicting Level 3 for Level 2) is more acceptable than a large deviation. We define WACC to give partial credit ($\alpha = 0.5$) for adjacent predictions: $ \text{WACC} = \frac{1}{N} \sum_{i=1}^{N} \left[ \mathbb{I}(d=0) + \alpha \cdot \mathbb{I}(d=1) \right]$, where $d = |\hat{y}_i - y_i|$.

   \item \textbf{Competency assessment} : We use standard classification accuracy (ACC).

\end{itemize}

\begin{table*}[!htbp]
  \centering
  \caption{Main performance comparison among ProIQA, LLM-based zero-shot evaluators, and supervised baselines across concept, difficulty, and competency assessment tasks. ``--'' indicates that the method is not applicable. The best results are highlighted in \textbf{bold}, and the second-best results are \underline{underlined}.}

  \begin{adjustbox}{max width=0.9\textwidth} 
\begin{tabular}{cc|cc|ccc|cc}
\toprule
\multicolumn{2}{c|}{Task} & \multicolumn{2}{c|}{(a) Concept} & \multicolumn{3}{c|}{(b) Difficulty} & \multicolumn{2}{c}{(c) Competency} \\
\multicolumn{2}{c|}{Dataset} & XES-500 & XES-1600 & Algebra & Algebra & XES-1500 & TIMSS19 & TIMSS23 \\
\multicolumn{2}{c|}{Metric} &  F1-Score &  F1-Score & Level-ACC & Level-WACC &  Pair-ACC & ACC & ACC \\
\midrule
GPT-4o & \multirow{3}[2]{*}{Closed source} & 0.44±0.04 & 0.28±0.02 & 0.27±0.02 & 0.54±0.01 & 0.29±0.02 & 0.41±0.03 & 0.40±0.04 \\
Qwen3-max &   & 0.51±0.04 & 0.36±0.02 & 0.30±0.01 & 0.60±0.01 & 0.30±0.02 & 0.46±0.04 & 0.32±0.05 \\
Gemini-2.5-pro &   & 0.50±0.02 & 0.32±0.02 & 0.23±0.01 & 0.46±0.02 & 0.26±0.01 & 0.55±0.05 & 0.50±0.06 \\
\midrule
Deepseek-R1 & \multirow{7}[2]{*}{Open source} & 0.53±0.02 & 0.35±0.01 & 0.24±0.01 & 0.47±0.02 & 0.26±0.02 & 0.36±0.05 & 0.35±0.05 \\
Qwen3-235B &   & 0.48±0.04 & 0.31±0.02 & 0.24±0.02 & 0.50±0.01 & 0.26±0.01 & 0.45±0.05 & 0.45±0.04 \\
Llama-3.3-70B &   & 0.41±0.03 & 0.26±0.01 & 0.31±0.02 & 0.61±0.01 & 0.34±0.02 & 0.37±0.05 & 0.33±0.05 \\
Qwen3-8B &   & 0.44±0.02 & 0.24±0.01 & 0.33±0.02 & 0.62±0.01 & 0.32±0.01 & 0.39±0.02 & 0.35±0.08 \\
Llama-3.1-8B &   & 0.23±0.02 & 0.09±0.00 & 0.20±0.02 & 0.44±0.02 & 0.28±0.01 & 0.37±0.07 & 0.37±0.05 \\
Mistral-7B-Instruct-v0.3 &   & 0.16±0.02 & 0.09±0.01 & 0.31±0.03 & 0.59±0.02 & 0.33±0.02 & 0.31±0.06 & 0.34±0.06 \\
\midrule
T-IRT & \multicolumn{1}{c|}{\multirow{6}[2]{*}{Supervised}} & \underline{0.70±0.04} & 0.65±0.01 & \underline{0.50±0.03} & \underline{0.73±0.02} & \underline{0.45±0.01} & — & — \\
R2DE &   & 0.41±0.04 & 0.43±0.01 & 0.42±0.02 & 0.67±0.02 & 0.37±0.02 & — & — \\
Qwen3-8B (SFT) &   & 0.66±0.03 & 0.53±0.02 & 0.35±0.03 & 0.63±0.01 & 0.34±0.02 & 0.49±0.03 & 0.43±0.07 \\
Llama-3.1-8B-Instruct(SFT) &   & 0.63±0.01 & 0.57±0.02 & 0.38±0.05 & 0.69±0.02 & 0.33±0.02 & 0.49±0.04 & 0.51±0.11 \\
Mistral-7B-Instruct-v0.3(SFT) &   & \underline{0.70±0.04} & \underline{0.70±0.02} & 0.39±0.06 & 0.67±0.07 & 0.33±0.02 & \underline{0.61±0.04} & \underline{0.58±0.08} \\
\midrule
ProIQA &   & \textbf{0.75±0.03} & \textbf{0.75±0.02} & \textbf{0.57±0.02} & \textbf{0.81±0.01} & \textbf{0.49±0.03} & \textbf{0.73±0.04} & \textbf{0.71±0.09} \\
\bottomrule
\end{tabular}%

\end{adjustbox}
\label{tab:Three task results}
\end{table*}

\paragraph{Results and Analysis}
Table~\ref{tab:Three task results} presents the comprehensive performance comparison across all three assessment tasks. We analyze each dimension below.

\begin{itemize}
    \item \textbf{Concept assessment} : As shown in Table~\ref{tab:Three task results}(a), ProIQA consistently achieves the highest F1 scores across both datasets, significantly outperforming the second-best baselines. The superiority of ProIQA can be attributed to its dual-view design. Traditional supervised methods (T-IRT/R2DE) rely solely on surface text, often missing implicit concept connections. Zero-shot LLMs, while knowledgeable, lack the specific alignment with the dataset's predefined concept taxonomy. In contrast, ProIQA leverages the reasoning tree to expose the logical steps where specific concepts are applied, enabling precise alignment with the target knowledge labels.

    \item \textbf{Difficulty assessment} : As shown in Table~\ref{tab:Three task results}(b), ProIQA outperforms all baselines on both datasets. On Algebra, ProIQA achieves the highest WACC, indicating that even when it errs, its predictions are closer to the ground truth. This suggests that the reasoning tree effectively captures the complexity of solution steps, which is the core driver of difficulty. On XES-1500, the superior pairwise accuracy demonstrates that ProIQA can effectively discern relative difficulty differences, a capability crucial for adaptive item recommendation systems.

   \item \textbf{Competency assessment} : As shown in Table~\ref{tab:Three task results}(c), the process-only variant of ProIQA (without reasoning-tree construction and GNN encoding, as TIMSS lacks item stems) achieves state-of-the-art accuracy, confirming that the complexity and logic in the solution process alone suffice for assessing higher-order competencies.

\end{itemize}

\subsection{To RQ2: Ablation Study}

To validate the contribution of our core components, we compared ProIQA variants: (1) \textbf{w/o Process:} Using only item stem representations. (2) \textbf{w/o Stream:} Using a linear solution text instead of a structured reasoning tree. (3) \textbf{w/o Graph:} Uses the textual representation of the entire reasoning tree without graph encoding. (4) \textbf{w/o GNN:} Performs pooling operations on each node, ignoring tree structure dependencies. (5) \textbf{ProIQA (Full):} Using the constructed reasoning tree. 

\begin{table}[t]
\centering
\caption{Ablation study of ProIQA components on concept and difficulty assessment tasks.}
\label{tab:Ablation Study}
\resizebox{0.47\textwidth}{!}{%
\begin{tabular}{c|w{c}{0.9cm}w{c}{0.9cm}|w{c}{1.1cm}w{c}{1.1cm}w{c}{1.1cm}}
\toprule
\textbf{Task } & \multicolumn{2}{c|}{\textbf{(a) Concept }} & \multicolumn{3}{c}{\textbf{(b) Difficulty }} \\
\textbf{Dataset } & \textbf{XES-500 } & \textbf{XES-1600 } & \multicolumn{2}{c}{\textbf{Algebra}} & \textbf{XES-1500 } \\
\textbf{Metric } & \textbf{F1-Score } & \textbf{F1-Score } & \textbf{Level-ACC } & \textbf{Level-WACC } & \textbf{Pair-ACC } \\
\midrule
\textbf{w/o Process} & 0.72±0.04 & 0.72±0.02 & 0.53±0.02 & 0.77±0.01 & 0.44±0.02 \\
\textbf{w/o Stream} & 0.73±0.04 & 0.73±0.01 & 0.55±0.02 & 0.79±0.01 & 0.46±0.03 \\
\textbf{w/o Graph} & 0.75±0.02 & 0.72±0.03 & 0.53±0.03 & 0.78±0.01 & 0.47±0.02 \\
\textbf{w/o GNN} & 0.73±0.03 & 0.73±0.02 & 0.52±0.01 & 0.78±0.00 & 0.47±0.02 \\
\textbf{ProIQA} & \textbf{0.75±0.03} & \textbf{0.75±0.02} & \textbf{0.57±0.02} & \textbf{0.81±0.01} & \textbf{0.49±0.03} \\
\bottomrule
\end{tabular}%
}
\end{table}

We present the comparative results of the ablation experiments in Table~\ref{tab:Ablation Study}. \textbf{Observation:} The ``w/o Process'' variant yields the lowest performance, confirming that surface stem alone is insufficient for fine-grained IQA — the answer process encodes deeper semantic meaning essential to the task. \textbf{Structure Matters:} ProIQA (Tree-based) outperforms ``w/o Stream'' (linear solution), demonstrating that the hierarchical reasoning tree, which separates independent subproblems from dependent logic, captures logical relationships more discriminatively than unstructured solution text. \textbf{Representation Quality:} ``w/o Graph'', despite employing a structured reasoning tree, degrades noticeably because its naive full-text representation introduces noise rather than extracting core process information. \textbf{Interaction Modeling:} ``w/o GNN'', which collapses node interactions via pooling, underperforms ProIQA, confirming that explicit modeling of inter-node relationships is necessary to integrate global reasoning tree signals effectively.

\subsection{To RQ3: Stability Analysis}

We investigated the sensitivity of ProIQA to two design choices central to its process-representation pipeline: the depth of the GNN encoder and the choice of the base embedding model. For this analysis we adopt the concept assessment task on the XES-500 dataset as a representative case, given its balanced scale and well-defined concept taxonomy.

\begin{figure}[t]
    \centering
    \begin{minipage}{\columnwidth}
        \centering
        \includegraphics[page=1,width=0.7\columnwidth]{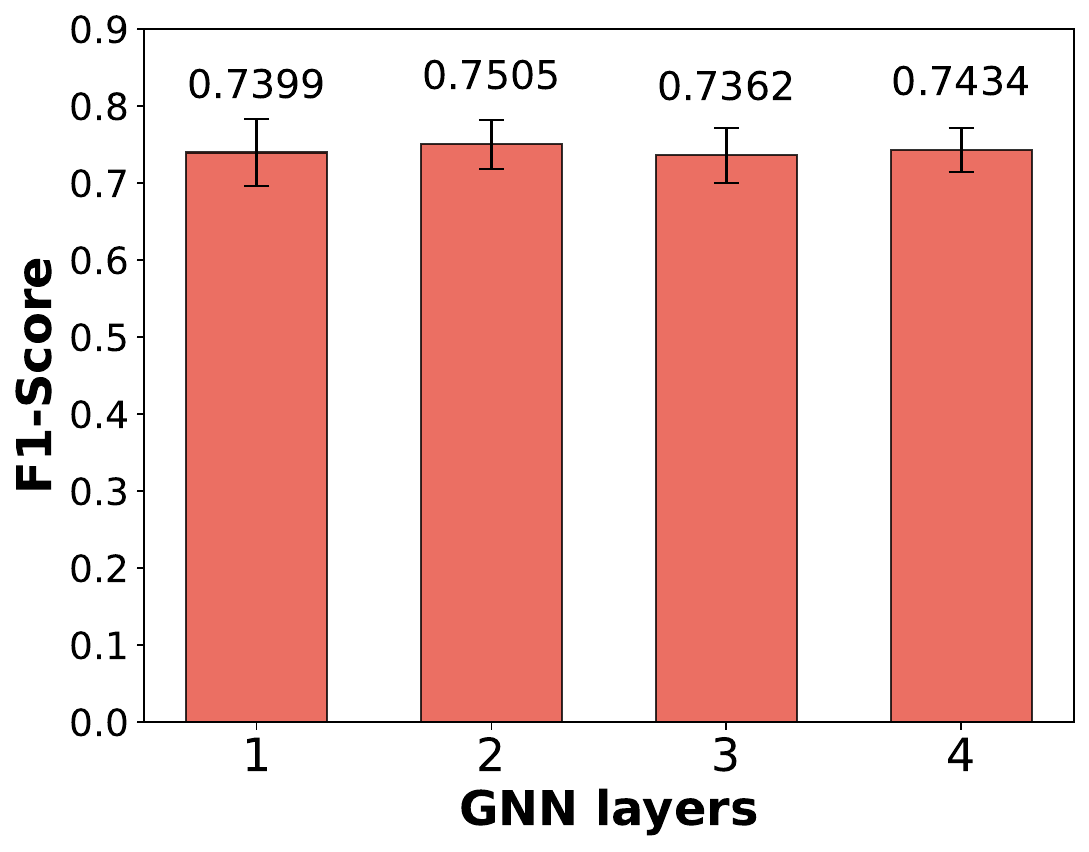}
        \caption{Effect of GNN depth on ProIQA's F1-score for concept assessment on XES-500.}
        \label{fig:Impact of GNN on F1}
    \end{minipage}
    
    \vspace{6pt}
    
    \begin{minipage}{\columnwidth}
        \centering
        \includegraphics[page=1,width=0.7\columnwidth]{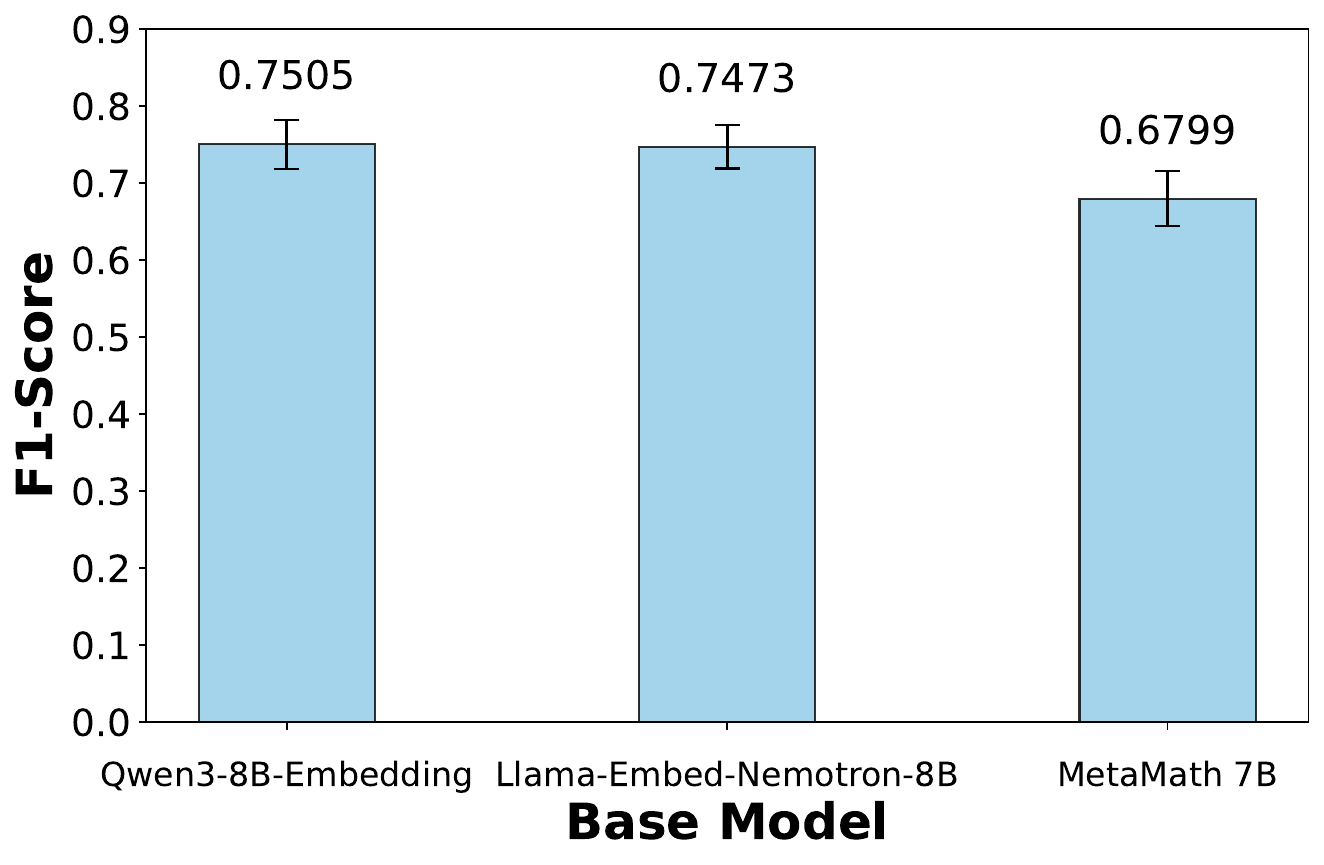}
        \caption{Effect of base embedding models on ProIQA's F1-score for concept assessment on XES-500.}
        \label{fig:Impact of base model on F1}
    \end{minipage}
\end{figure}

\textbf{Impact of GNN Layers:} Figure~\ref{fig:Impact of GNN on F1} shows that F1 scores remain stable across 1 to 4 layers, with a maximum deviation of only 1.5\% from the mean. This flat profile suggests that ProIQA is data-efficient and does not require extremely deep networks to capture reasoning dependencies, making it deployable in resource-constrained environments.

\textbf{Impact of Base Model:} Figure~\ref{fig:Impact of base model on F1} compares specialized embedding models (Qwen3-Embed, Llama-Embed) against a general math model (MetaMath-7B). Results indicate that embedding-optimized models yield better performance. This gap indicates that general math models, though competent at mathematical generation, produce less discriminative vector representations when used as frozen encoders. Notably, the two specialized embedding models yield near-identical performance despite their different training paradigms, suggesting that ProIQA's dual-view architecture is robust to the choice of embedding backbone as long as it is optimized for semantic representation. Thus, any modern embedding model can serve as a drop-in encoder without substantial performance loss.

\subsection{To RQ4: Three-dimensional applicability}

To concretely demonstrate ProIQA's advantage in fine-grained, multi-dimensional assessment, we select a representative item pair for detailed analysis. Item A asks students to determine the duration of a power outage given two candles with different burn rates ($6h$ and $10h$) and a final condition relating their remaining lengths. Item B asks for the side length and area of a square given its $24$ cm perimeter.

These two items originally carried ground truth labels for knowledge concepts and difficulty, but lacked competency annotations. To enable a unified three-dimensional evaluation, we invited human experts to annotate their disciplinary competencies, thereby establishing complete ground truth across all three dimensions for this pair. We then compare ProIQA predictions against those of a representative LLM baseline (Deepseek-V3.2, zero-shot) on the same pair. The results are summarized as follows.

\textbf{Knowledge concept.} The ground truth assigns Item A to the ``Varying-Multiple Problem'' category, and Item B under ``Geometric Cognition''. ProIQA correctly predicts both assignments. The LLM baseline misclassifies Item A as a ``Linear Equation Word Problem'' and reduces Item B to superficial descriptors such as ``Perimeter'' and ``Formula Application'', failing to capture higher-level conceptual categories.

\textbf{Disciplinary competency.} Human experts annotate Item A as ``Knowing'' (the solution follows a standard varying-multiple template requiring only routine pattern recognition) and Item B as ``Reasoning'' (deriving side length from perimeter and subsequently computing area demands genuine geometric understanding that goes beyond formula recall). ProIQA matches both truth labels exactly. The LLM baseline produces the complete opposite: it labels Item A as ``Reasoning'', deceived by the lengthy narrative and the presence of algebraic variables, and labels Item B as ``Knowing'', misled by its textual brevity.

\textbf{Difficulty.} The ground truth establishes that Item B is harder than Item A ($B>A$), consistent with its higher competency demand. ProIQA correctly predicts this relative ordering. The LLM baseline, conflating textual verbosity with cognitive depth, predicts the reverse ($A > B$).

In summary, on this item pair, ProIQA achieves exact alignment with ground truth across all three dimensions, whereas the LLM baseline misclassifies concepts, reversing competency levels, and inverting the difficulty ordering. This case clearly illustrates the core weakness of stem-only assessment: without modeling the solution process, surface-level methods systematically confuse how an item looks with what it actually demands. ProIQA's dual-view paradigm resolves this ambiguity by jointly reasoning over the item stem and its structured solving path.

\subsection{Qualitative Error Analysis}

To complement the quantitative results, we conduct a qualitative analysis of prediction-label disagreements to identify systematic error patterns and failure modes. We focus on concept and difficulty assessment, where the reasoning tree plays a central role.

\textbf{Concept Assessment.}
Two primary error patterns emerge. First, \textbf{Semantic Similarity}: the top-K confused concept pairs (e.g., ``Sum-Multiple'' vs.\ ``Difference-Multiple Problems,'' or standard vs.\ transformed ``Profit-Loss Problems'') consistently fall within the same macro-category, indicating that ProIQA reliably captures the hierarchical knowledge structure despite occasional drift between closely related sub-concepts. Second, \textbf{Multiple Valid Solution Paths}: manual audit reveals that the LLM may generate an alternative yet mathematically valid reasoning path that maps to a concept different from the ground-truth label. For instance, a ``Chickens-and-Rabbits'' problem labeled ``Grouping Method'' was solved via ``Assumption Method'' in the reasoning tree. Such discrepancies reflect inherent solution diversity rather than framework failure.

\textbf{Difficulty Assessment.}
Three patterns are observed. First, \textbf{Adjacent-Level Concentration}: consistent with the high WACC reported in Table~\ref{tab:Three task results}, errors predominantly land on neighboring levels, confirming that the ordinal difficulty structure is preserved even in misclassifications. Second, \textbf{Annotation Noise}: some apparent errors stem from inconsistent benchmark labels (e.g., a completing-the-square problem labeled Level~4 was rated low difficulty by ProIQA, a judgment more consistent with its pedagogical demand). 
Third, Pseudo-Hierarchical Reasoning Trees: a flattened tree under-represents cognitive depth. For example, Algebra item 341 (Decimal Construction, Level 5) had its full solution placed in the root, with children merely restating root segments, causing ProIQA to predict Level 2. This shows that a logically correct yet structurally shallow tree offers limited signal for distinguishing deep multi-step reasoning from routine computation.

\subsection{Time Complexity Analysis}

We analyze the time complexity of ProIQA and compare it with two supervised baselines, T-IRT and R2DE. Item quality assessment is inherently offline, so one-time tree construction and verification do not affect inference latency. Taking the concept assessment task as a representative case, the dominant computational cost in ProIQA comes from the GNN encoder; the final classifier consists of shallow linear transformations and is negligible in comparison. The GNN aggregates node features over $L_{\text{gnn}}$ layers to produce a global graph representation, yielding a forward-pass complexity of $O(N|\mathcal{C}|L_{\text{gnn}}d^2)$, where $N$ is the number of samples, $|\mathcal{C}|$ is the number of target concepts, and $d$ is the PLM hidden dimension. Including backpropagation, the overall training complexity is $O(N|\mathcal{C}|L_{\text{gnn}}|d^2)$.

For T-IRT, the BERT backbone is fine-tuned end-to-end during training, which embeds the semantic extraction process into the optimization loop. Its self-attention layers contribute a complexity of $O(N|L|d^2|l)$, where $L$ is the number of Transformer layers and $l$ is the average input sequence length. With backpropagation, this becomes $O(N|L|d^2|l)$. R2DE employs TF-IDF feature extraction followed by a random forest classifier, yielding a complexity of $O(|\mathcal{C}|T|N|F|D|\log(N))$, where $T$ is the number of decision trees, $F$ is the TF-IDF feature dimension, and $D$ is the maximum tree depth.

T-IRT and ProIQA exhibit comparable complexity, both substantially lower than R2DE, whose many decision trees over high-dimensional TF-IDF features yield a large multiplicative constant. ProIQA thus achieves strong performance relative to traditional methods without substantial additional overhead, confirming that the dual-view architecture attains its accuracy gains at manageable cost.

\section{Discussion}
\textbf{Generality Beyond Mathematics.} Although this work focuses on K12 mathematics, ProIQA's process-representation paradigm is domain-agnostic: any subject whose assessment items admit decomposable solution paths, such as physics, chemistry, biology, programming and beyond can be represented as reasoning trees and evaluated through the same dual-view architecture. Extending ProIQA to these domains and to additional quality dimensions such as item clarity and normativity is a natural direction for future work.

\textbf{On Difficulty Versus Discrimination.} IRT calibration of XES3G5M yields both a difficulty and a discrimination parameter per item. We use only the former. Difficulty reflects the cognitive demand of an item — the concepts involved, the reasoning steps required — and is stable across calibration samples. Discrimination captures how sharply an item separates examinees of nearby ability. Since XES3G5M is drawn from a single-grade, single-platform cohort, its discrimination estimates are population-specific and less reliable as general-purpose training labels for a content-based IQA model. Difficulty, by contrast, is invariant across populations and reflects an intrinsic item property, so we adopt it as the sole label in our task formulation.

\textbf{Limitations and Future Work.} 
First, automated verification ensures over 90\% of reasoning trees are logically coherent, yet does not guarantee their pedagogical representativeness, which expert evaluation would help establish. Second, the tree-structured representation may not capture reasoning patterns involving cross-branch analogy or iterative refinement; exploring richer process structures is worth pursuing. Third, the frozen pre-trained language may introduce a gap between extracted and task-optimal representations, incorporating the PLM into the training pipeline bridge this gap. Fourth, without item stems in TIMSS, competency assessment validates only ProIQA's process view. Finally, our small subsets and IRT-estimated labels call for larger, expert-annotated benchmarks. 

\section*{ACKNOWLEDGMENTS}
We would like to thank the anonymous reviewers for their constructive comments. This work is supported by the National Natural Science Foundation of China (No. 62476091) and Fundamental Research Funds for the Central Universities (Grant No. 2026ECNU-WLJC009). The corresponding author of this work is Hong Qian.

\bibliographystyle{IEEEtran}
\bibliography{references}





\end{document}